\documentclass[]{wam}

\usepackage[utf8]{inputenc}
\usepackage[T1]{fontenc}
\usepackage{url}
\usepackage{booktabs}
\usepackage{amsfonts}
\usepackage{amsmath}
\usepackage{amssymb}
\usepackage{microtype}
\usepackage{textcomp}
\usepackage{graphicx}
\usepackage{array}
\usepackage{multirow}
\usepackage{makecell}
\usepackage{paralist}
\usepackage{wrapfig}
\usepackage{colortbl}
\usepackage{xspace}

\graphicspath{{./}{figures/}}

\definecolor{oursgreen}{RGB}{230,245,235}
\newcommand{\best}[1]{\cellcolor{oursgreen}\textbf{#1}}
\newcommand{\etal}{et~al.}

\providecommand{\keywords}[1]{\noindent\textbf{Keywords:} #1}

\title{GraspFoM: Towards Reconstruction-Driven Robotic Grasping with 3D Foundation Priors}

\author[1,*]{Dongli Wu}
\author[2,*]{Xiaobao Wei}
\author[2]{Hao Wang}
\author[2]{Qiaochu Dong}
\author[2,3]{Ying Li}
\author[2]{Qingpo Wuwu}
\author[2,\dag]{Ming Lu}
\author[1,\ddag]{Wufan Zhao}

\affiliation[1]{The Hong Kong University of Science and Technology (Guangzhou)}
\affiliation[2]{Peking University}
\affiliation[3]{The Hong Kong University of Science and Technology}
\contribution[*]{Equal contribution}
\contribution[\dag]{Project leader}
\contribution[\ddag]{Corresponding author}

\abstract{
Robotic grasping is a fundamental capability in robotic manipulation. 
Yet grasping remains challenging under partial observations. 
Reliable grasping depends on both local contact cues and object-level 3D structure. 
Existing geometry-aware grasping methods recognize the value of reconstruction, but they typically treat geometry as an intermediate prediction rather than a reusable object prior for grasping. 
In this paper, we present GraspFoM, a unified framework that leverages 3D foundation priors (SAM3D) to build a shared 3D object latent for both reconstruction and grasp pose prediction. 
Built on this shared object latent, we introduce an anchor-initialized truncated pose-reasoning diffuser that predicts continuous and multimodal grasp poses without directly relying on discrete grasp candidates. 
We further investigate the interaction between reconstruction and grasping through a reconstruction-aware scorer and a residual latent updater. Reconstruction provides grounded geometric cues, while grasp supervision refines the shared object latent toward grasp-relevant affordances. 
GraspFoM jointly predicts grasp poses and reconstructs high-fidelity 3D assets in mesh and 3DGS forms. 
Comprehensive experiments demonstrate that GraspFoM achieves state-of-the-art results on both reconstruction and grasping. 
Notably, these improvements require only a small number of additional trainable parameters. Component-wise ablation studies also demonstrate the contribution of each component. 
Codes will be released. 

}

\date{\today}
\code{\href{https://annike-val.github.io/GraspFoM/}{https://annike-val.github.io/GraspFoM/}}

\begin{document}
\maketitle


\section{Introduction}

Robotic grasping is a fundamental capability in robotic manipulation, yet it remains challenging in real-world scenes~\cite{du2021vision, dong2023review, newbury2023deep,wang2025roboarmgs}. A robust grasping system~\cite{xu2023joint, deng2025graspvla, li2024clickdiff,li2026manipdreamer3d} must jointly reason about object geometry and multimodal action policies under clutter, occlusion, and partial observations. This makes accurate and generalizable grasping difficult in practice. 

\begin{figure}[hbt]
  \centering
  \includegraphics[width=\textwidth]{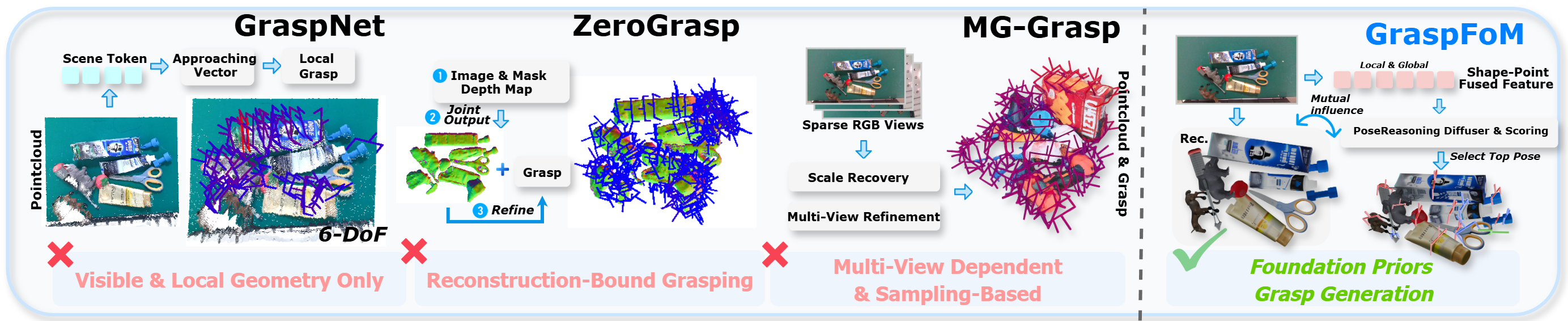}
    \caption{Representative paradigms for robotic grasping. GraspNet~\cite{fang2020graspnet} relies on visible local geometry, ZeroGrasp~\cite{iwase2025zerograsp} couples grasping with task-specific reconstruction, and MG-Grasp~\cite{wang2026mg} depends on sparse multi-view sampling and refinement. In contrast, GraspFoM builds grasping on transferable 3D foundation priors, using a shared object latent to unify reconstruction, continuous grasp generation, and reconstruction-aware scoring.}
  \label{fig:teaser}
\end{figure}

Existing grasping methods have evolved along several major directions as shown in Fig.~\ref{fig:teaser}. Early learning-based approaches~\cite{fang2020graspnet,mousavian20196,wang2021graspness} mainly formulate grasping as direct pose prediction from RGB-D observations or point clouds, establishing the basic paradigm of data-driven grasp detection. Building on this line, subsequent works~\cite{sundermeyer2021contact,zhou2024you} increasingly incorporate richer geometric cues, such as contact reasoning, graspness estimation, and object-centric 3D features, which improve grasp generation in cluttered scenes and for novel objects. 
More recently, several works~\cite{chisari2024centergrasp,fan2025neugrasp,iwase2025zerograsp, wang2026mg} have moved toward unified frameworks that couple object reconstruction with grasp pose prediction, bringing explicit 3D supervision into grasp learning. In parallel, Vision-language-model (VLM) based policies~\cite{shao2025large,deng2025graspvla,cao2026fastdrivevla,cao2026evodrivevla} further push grasping toward broader task-agnostic generalization by leveraging the strong semantic reasoning, prior knowledge, and instruction-following capabilities inherited from large language models. These methods highlight the growing role of large-scale foundational priors in improving the generalization ability of robotic grasping. 

Despite these advances, robotic grasping under partial observations remains challenging because the underlying object structure is often incomplete or ambiguous. 
Reconstruction-aided grasping improves geometric grounding, but existing methods~\cite{chisari2024centergrasp,fan2025neugrasp, iwase2025zerograsp} still rely on task-specific representations learned from scratch. 
By contrast, VLM-based policies~\cite{shao2025large,deng2025graspvla, noh2025graspsam} improve task-level generalization, yet they inherit 2D semantic knowledge and rely on time-consuming LLM decoding, leading to poor and slow 3D grasping pose prediction. 

Meanwhile, general 3D object reconstruction has progressed rapidly from task-specific geometric regression~\cite{wang2021neus, yariv2021volume, cen2023segment, wei2024nto3d, huang2024s3g} and optimization pipelines to large-scale reconstruction~\cite{tochilkin2024triposr, yang2024gaussianobject, xu2024instantmesh, wei2025gazegaussian, wu2025dnrselect} and generation~\cite{qian2023magic123, shi2023zero123++, deitke2023objaverse, zeng2025rethinking, chen2025robotwin} models that learn reusable shape priors from massive data. Recent methods~\cite{hong2023lrm, zhang2024gs, chen2025sam} such as LRM, GS-LRM and SAM3D show that high-quality 3D geometry can now be recovered in a feed-forward manner from sparse or incomplete observations, producing faithful object assets in forms such as meshes, radiance fields~\cite{mildenhall2021nerf}, and 3D Gaussians~\cite{kerbl20233dgs}. 
These advances suggest that strong object-level 3D priors can be learned and transferred across novel instances and observation conditions. Yet existing methods mainly optimize the fidelity of digital 3D assets, rather than the object representations needed for robotic manipulation. Consequently, their priors cannot be directly applied to grasping, which requires not only accurate geometry but also manipulation-relevant structure such as graspable regions and affordance cues~\cite{pohl2020affordance, zeng2022robotic, zhong2025dexgrasp, wei2025afforddexgrasp}.

Motivated by this gap, we propose GraspFoM, a unified framework for reconstruction-driven robotic grasping built on top of SAM3D~\cite{yang2023sam3d} priors. Given RGB images, object masks, and point maps, GraspFoM first uses the sparse-structure stage of SAM3D to initialize a shared 3D object latent. 
Specifically, we query surface points from the object and fuse point-wise local cues with grid-sampled global shape context to construct shared 3D object latents. 
Based on these latents, we introduce a pose-reasoning diffuser initialized with an anchored Gaussian distribution, which models continuous and multimodal grasp pose distributions through a reasoning trunk and a truncated denoising process.
Unlike existing methods that rely on discrete grasp candidates, our method predicts grasp poses directly from the shared object latents. 

Beyond joint prediction, GraspFoM enforces bidirectional interaction between reconstruction and grasp learning through a reconstruction-aware scorer and a residual latent updater.  
The scorer predicts graspness, quality, and affordance directly on reconstructed geometry, grounding grasp evaluation in object structure rather than pose-only cues. The updater takes grasp supervision back into the shared latent through surface-centered offsets, injecting manipulation-relevant signals into the reconstruction process. This establishes a bidirectional link between the two tasks: reconstruction improves grasp prediction, while grasp learning refines the underlying object representation. As a result, GraspFoM jointly outputs accurate grasp poses along with high-fidelity 3D assets, achieving state-of-the-art performance on the widely used GraspNet-1B~\cite{fang2020graspnet} benchmark. 

Our contributions are summarized as follows:
\begin{compactitem}
    \item We propose GraspFoM, a unified framework that leverages 3D foundation priors to build a shared 3D object latent for both object reconstruction and grasp pose prediction.
    \item We introduce an anchor-initialized pose-reasoning diffuser as the grasping head to model continuous 3D grasp pose distributions, thereby avoiding discrete candidates. 
    \item We design a reconstruction-aware scorer and a residual latent updater to couple reconstruction and grasp learning, enabling geometric grounding and grasp-relevant affordance enhancement within the shared latent. 
    \item GraspFoM achieves state-of-the-art performance on GraspNet-1B dataset. Comprehensive experiments demonstrate the effectiveness of proposed techniques. 
\end{compactitem}

\section{Related Work}
\subsection{3D Object Reconstruction}
3D object reconstruction aims to recover faithful geometry and appearance from sparse observations. Early works~\cite{wang2021neus,yariv2021volume,cen2023segment,wei2024nto3d,huang2024s3g,wang2025plgs} mainly rely on implicit neural representations, signed distance fields, or optimization-based formulations to model object surfaces with high fidelity. 
Recent progress~\cite{tochilkin2024triposr,yang2024gaussianobject,xu2024instantmesh,wei2025gazegaussian,wu2025dnrselect,qian2023magic123,shi2023zero123++,deitke2023objaverse,zeng2025rethinking,chen2025robotwin} shifts toward scalable reconstruction and generation models trained on large 3D corpora, enabling feed-forward recovery of object geometry from sparse or incomplete observations. In particular, large reconstruction models such as LRM~\cite{hong2023lrm}, GS-LRM~\cite{zhang2024gs}, and SAM3D~\cite{chen2025sam,yang2023sam3d} show that strong object-level priors can be learned from massive data and transferred across novel instances and observation conditions. These methods can produce high-quality 3D assets in forms such as meshes, radiance fields~\cite{mildenhall2021nerf}, and 3D Gaussians~\cite{kerbl20233dgs}. However, existing reconstruction methods mainly optimize the fidelity of digital assets rather than the manipulation-oriented object structure required for robotic interaction. This leaves a gap between recent advances in general 3D reconstruction and the needs of robotic manipulation.

\subsection{Robotic Reconstruction for Grasping}
Accurate 3D geometric representation is foundational to robust robotic grasping~\cite{williams2024fvdb}. Traditional approaches rely on discrete formats (e.g., voxel grids and point clouds) to model object shapes. While widely adopted in early grasp pipelines, these methods often trade resolution for efficiency or lack generalizability across diverse geometries~\cite{wang2023octree,williams2024fvdb,eppner2021acronym,park2019deepsdf}.
For cluttered real-world environments, recent methods emphasize open-world adaptability. For instance, SceneComplete~\cite{agarwal2024scenecomplete} fills missing geometries via 3D scene completion, and SAM-based approaches~\cite{qu2024sam} leverage powerful segmentation models to reconstruct unseen objects. 
Beyond traditional discrete formats, recent advances in continuous 3D representations, such as neural occupancy fields and 3D Gaussians~\cite{wei2024nto3d, li2025manipdreamer, li2026manipdreamer3d, wang2025embodiedocc++, wei2025omniindoor3d}, have significantly enhanced embodied spatial perception, scene understanding, and trajectory planning. 
Although these works highlight the critical role of complete 3D modeling, they focus primarily on scene understanding or spatial navigation without explicit integration into end-to-end grasp reasoning. 
Similarly, while implicit encoders enhance shape details~\cite{huang2023neural,heppert2023carto,chan2023generative}, and fast reconstruction frameworks prioritize speed for immediate grasp planning~\cite{avigal20206}, their task-specific designs or limited precision restrict their adaptability to dynamic, open-world scenarios. Furthermore, recent advancements like Sharp-It~\cite{edelstein2025sharp} refine low-fidelity 3D geometry into high-precision multi-view representations via diffusion models, but they still lack synergy with downstream grasp optimization.

\subsection{Grasping Pose Prediction} 
Early grasp pose prediction methods typically output 6D poses directly from RGB-D inputs, often bypassing explicit geometric modeling, which leads to unstable contacts or collisions in cluttered scenes~\cite{fang2020graspnet,mousavian20196,wang2021graspness}. To address these flaws, subsequent works have advanced grasping in targeted scenarios. For instance, Contact-GraspNet~\cite{sundermeyer2021contact} enables efficient 6-DoF grasp generation in clutter, while recent dynamic scene reconstruction pipelines~\cite{zhou2024you} facilitate the grasping of novel objects. However, these methods often rely on task-specific designs and lack generalizable 3D perception support.
To tightly couple perception and planning, unified reconstruction-grasp frameworks attempt to co-optimize geometric modeling and pose prediction. For example, CenterGrasp~\cite{chisari2024centergrasp} learns object-aware implicit representations for simultaneous shape reconstruction and grasp estimation. Nevertheless, such approaches face inherent trade-offs among reconstruction detail, inference speed, and cross-scene adaptability~\cite{ma2024generalizing,liu2023meshdiffusion,chan2023generative,chisari2024centergrasp}. Recent large-scale efforts, such as ZeroGrasp~\cite{iwase2025zerograsp}, leverage massive datasets (e.g., 11.3B annotations) to achieve zero-shot grasping via unified octree-based reconstruction, but they remain susceptible to resolution dependency under occlusions. Concurrently, Vision-Language-Action (VLA) models~\cite{shao2025large} and their variants like GraspVLA~\cite{deng2025graspvla} have shown remarkable generalization in dynamic scenarios. Yet, these VLM-based approaches typically prioritize semantic generalization over the high-precision geometric alignment required for stable grasping. 

In contrast, we build GraspFoM on top of SAM3D priors and use them to form a shared 3D object latent for both reconstruction and grasp pose prediction. This design differs fundamentally from recent unified methods such as ZeroGrasp~\cite{iwase2025zerograsp}, which learns from scratch and improves grasping through reconstruction. By leveraging transferable 3D foundation priors, our method focuses on reusable object-level representation rather than task-specific geometry learning. 

\begin{figure*}[!ht]
  \includegraphics[width=\textwidth]{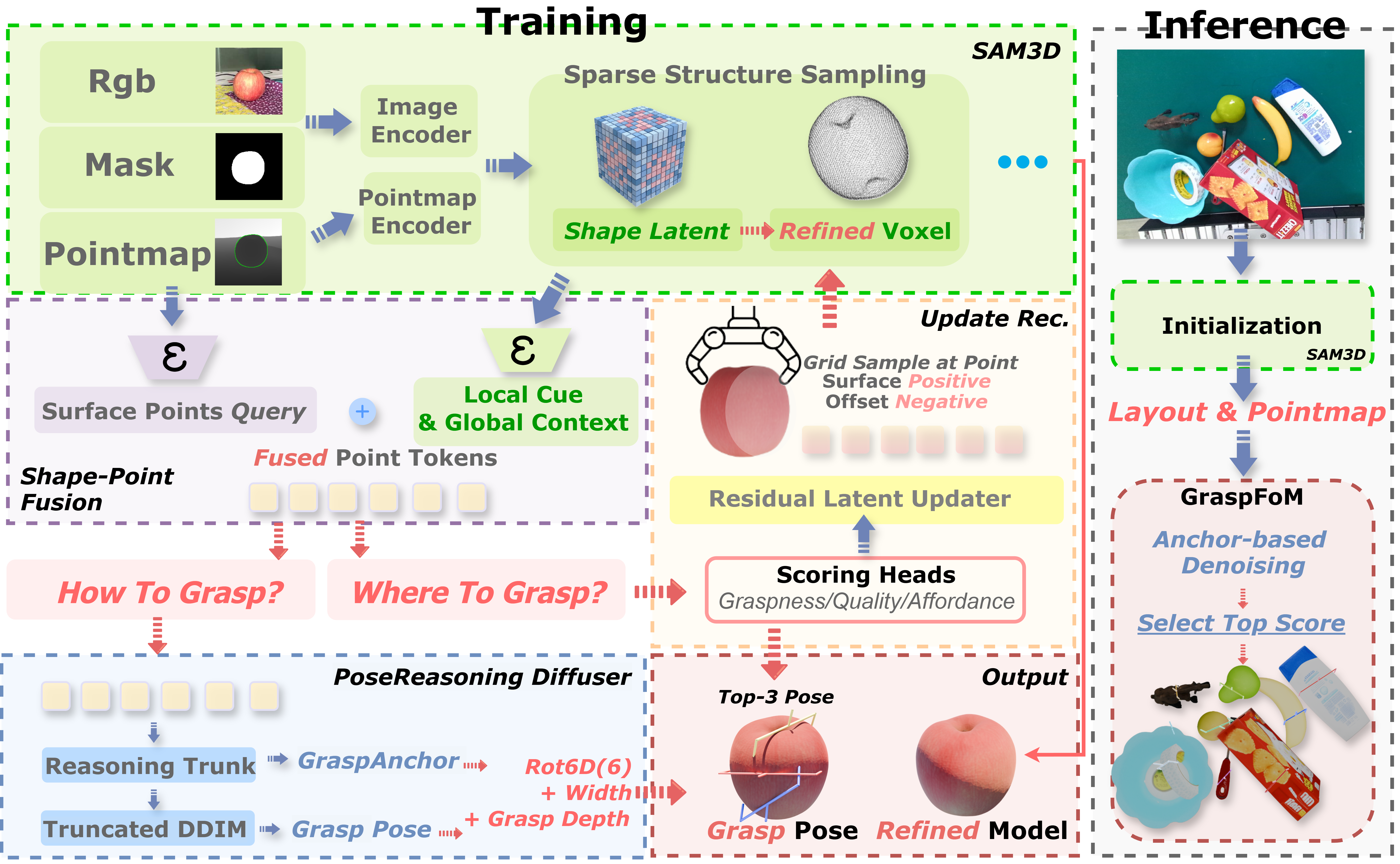}
  \vspace{-6mm}
    \caption{Overview of GraspFoM. (a) During training, we first fuse a shared shape latent. Based on these latents, the pose-reasoning diffuser predicts grasp poses and further refines the latents. (b) During inference, GraspFoM performs anchor-based denoising, score-based selection, and returns the top-ranked grasp poses together with high-fidelity 3D object assets.}
  \vspace{-4mm}
  \label{fig:pipeline}
\end{figure*}

\section{Methodology}

The overall pipeline of GraspFoM is illustrated in Fig.~\ref{fig:pipeline}. We first introduce the problem setup in Sec.~\ref{sec:preliminaries}. We then describe how 3D foundational priors are transformed into a shared object representation in Sec.~\ref{sec:shared_latent}, followed by the pose generation process based on anchor-conditioned denoising in Sec.~\ref{sec:diffuser}. The reconstruction-aware scoring and latent refinement are presented in Sec.~\ref{sec:scorer_reconstruction}. 

\subsection{Preliminaries}
\label{sec:preliminaries}

Let $\mathbf{I} \in \mathbb{R}^{H \times W \times 3}$ denote an RGB image, $\mathbf{M} \in \{0,1\}^{H \times W}$ the object mask, and $\mathbf{P} \in \mathbb{R}^{H \times W \times 3}$ the object-centric point map in the camera frame. Our goal is to jointly reconstruct an object-centric 3D representation and predict a set of feasible grasp poses from the same observation. We denote the queried visible surface points by
\begin{equation}
\mathcal{Q} = \{(\mathbf{p}_i, \mathbf{n}_i)\}_{i=1}^{N}, \qquad \mathbf{p}_i,\mathbf{n}_i \in \mathbb{R}^{3},
\end{equation}
where $\mathbf{p}_i$ is a visible 3D point and $\mathbf{n}_i$ is its corresponding surface normal.

We adopt the standard two-finger parallel gripper parameterization. For each queried point, a grasp pose is represented as
\begin{equation}
\mathbf{g}_i = [\mathbf{r}_i, w_i, d_i] \in \mathbb{R}^{8},
\end{equation}
where $\mathbf{r}_i \in \mathbb{R}^{6}$ is the continuous 6D rotation representation, $w_i$ is the finger opening width, and $d_i$ is the grasp depth. Following the implementation, the diffusion branch models only rotation and depth, while the width is recovered from local geometry during inference. We therefore define the diffusion target as
\begin{equation}
\mathbf{x}_i^{0} = [\bar{\mathbf{r}}_i, \bar{d}_i] \in [-1,1]^{7},
\end{equation}
where $\bar{\mathbf{r}}_i$ and $\bar{d}_i$ are normalized versions of the 6D rotation and depth, respectively.

In addition to pose prediction, our model predicts three point-wise scores,
\begin{equation}
s_i \in \mathbb{R}, \qquad q_i \in \mathbb{R}, \qquad a_i \in \mathbb{R},
\end{equation}
which correspond to graspness, grasp quality, and affordance. These scores are used both for training and for reconstruction-aware ranking at inference time.

\subsection{Shared Object Latent Preparation}
\label{sec:shared_latent}
A key challenge is that reconstruction and grasping require both fine-grained local geometry and coherent global object structure, yet existing pipelines usually encode them in separate task-specific representations. We therefore seek a shared object latent that inherits strong 3D priors from SAM3D while remaining directly usable for downstream grasp reasoning.

We build GraspFoM on top of the object-centric stage-1 output of SAM3D~\cite{yang2023sam3d}. Given $(\mathbf{I}, \mathbf{M}, \mathbf{P})$, the SAM3D sparse-structure pipeline produces a compact object latent together with the coordinate alignment parameters used by the point-map preprocessing:
\begin{equation}
(\mathbf{z}, \boldsymbol{\gamma}, \boldsymbol{\beta})=\Phi_{\text{SAM3D}}(\mathbf{I}, \mathbf{M}, \mathbf{P}),
\end{equation}
where $\mathbf{z}\in\mathbb{R}^{G^3\times C_z}$ denotes the shared 3D object latent, and $(\boldsymbol{\gamma}, \boldsymbol{\beta})$ denote the scale-shift parameters for aligning camera-space points with the latent coordinate system.

To make this compact latent directly usable for downstream grasp reasoning, we first convert it into a structured 3D feature volume and then lift it into a shape-aware grid representation:
\begin{equation}
\widetilde{\mathbf{Z}} = E_{\text{shape}}\!\left(\mathrm{Reshape}(\mathbf{z})\right)\in\mathbb{R}^{C\times G^3},
\end{equation}
where $E_{\text{shape}}$ denotes a lightweight 3D encoder shared by local feature sampling and global token extraction. This shape grid serves as the common geometric representation for both reconstruction and grasp prediction.

Given the shape-aware grid $\widetilde{\mathbf{Z}}$, we next build a latent-conditioned representation for each queried surface point $\mathbf{p}_i \in \mathcal{Q}$, where $\mathcal{Q}$ is obtained by the surface-point query module from the visible object surface. Specifically, each queried point is first transformed from the camera coordinate system to the latent-aligned coordinate system and then normalized to the sampling range of the latent grid:
\begin{equation}
\bar{\mathbf{p}}_i = \mathcal{N}\!\left(\mathcal{T}_{\text{ssi}}(\mathbf{p}_i;\boldsymbol{\gamma},\boldsymbol{\beta})\right),
\end{equation}
where $\mathcal{N}(\cdot)$ denotes normalization to $[-1,1]^3$.

Based on $\bar{\mathbf{p}}_i$, we extract local geometric cues and latent-conditioned shape features to form the point-wise local representation:
\begin{equation}
\mathbf{f}_i^{\text{pt}} = E_{\text{pt}}([\mathbf{p}_i,\mathbf{n}_i]), \qquad
\mathbf{f}_i^{\text{loc}} = E_{\text{loc}}\!\left(\mathcal{S}(\widetilde{\mathbf{Z}}, \bar{\mathbf{p}}_i)\right),
\end{equation}
where $\mathbf{n}_i$ is the normal of $\mathbf{p}_i$, and $\mathcal{S}(\cdot,\cdot)$ denotes trilinear sampling on the shared 3D latent grid. 

To inject global object context, we further convert the encoded latent grid into shape tokens $\mathbf{T}_{\text{shape}} = E_{\text{tok}}(\widetilde{\mathbf{Z}})$ and update each point feature through cross-attention. The resulting feature is fused with the local point and latent features to form the final fused point token:
\begin{equation}
\mathbf{f}_i =
E_{\text{fuse}}
\!\left[
\mathbf{f}_i^{\text{pt}};
\mathbf{f}_i^{\text{loc}};
\mathrm{MHA}\!\left(\mathbf{f}_i^{\text{pt}}+\mathbf{f}_i^{\text{loc}}, \mathbf{T}_{\text{shape}}, \mathbf{T}_{\text{shape}}\right)
\right].
\end{equation}
Those fused point tokens jointly encode local surface cues and global shape context, and serves as the shared representation for the subsequent pose-reasoning, scoring, and reconstruction-update branches. 

\subsection{Pose-reasoning Diffuser}
\label{sec:diffuser}

The grasp space is diverse and inherently multimodal. Direct regression tends to collapse to a single mode. We therefore formulate grasp prediction as latent-conditioned denoising in a normalized pose space. 

We first build an anchor set in diffusion space by clustering valid training poses:
\begin{equation}
\mathcal{A} = \{\mathbf{a}_k\}_{k=1}^{K}, \qquad \mathbf{a}_k \in [-1,1]^7.
\end{equation}
Each anchor is obtained by applying K-means to the normalized ground-truth pose targets over the training set. Given a training target $\mathbf{x}_i^0$, we assign it to the nearest anchor,
\begin{equation}
k_i^{\star} = \arg\min_{k} \|\mathbf{x}_i^0 - \mathbf{a}_k\|_2^2.
\end{equation}

Instead of diffusing from pure Gaussian noise, we adopt a truncated anchored diffusion process. For a randomly sampled timestep $t \in \{0, \dots, T_{\text{tr}}-1\}$ and Gaussian noise $\boldsymbol{\epsilon} \sim \mathcal{N}(\mathbf{0}, \mathbf{I})$, we construct the noisy state by
\begin{equation}
\mathbf{x}_i^t =
\sqrt{\bar{\alpha}_t}\,\mathbf{a}_{k_i^{\star}} +
\sqrt{1-\bar{\alpha}_t}\,\boldsymbol{\epsilon},
\end{equation}
where $\bar{\alpha}_t$ is the cumulative noise coefficient. This design follows the intuition of truncated diffusion: a more informative initialization improves denoising efficiency, while the anchor-conditioned formulation still supports multimodal grasp generation. 

Conditioned on the fused latent $\mathbf{f}_i$, the pose branch first computes a pose reasoning feature
\begin{equation}
\mathbf{h}_i = E_{\text{pose}}(\mathbf{f}_i),
\end{equation}
and predicts both the anchor logits and the denoised pose target:
\begin{equation}
\boldsymbol{\pi}_i = E_{\text{mode}}(\mathbf{h}_i) \in \mathbb{R}^{K},
\qquad
\hat{\mathbf{x}}_i^0 = D_{\theta}(\mathbf{h}_i, \mathbf{x}_i^t, t),
\end{equation}
where $\boldsymbol{\pi}_i$ denotes the unnormalized scores over $K$ grasp anchors, and $D_{\theta}$ is the pose-reasoning diffuser with sinusoidal timestep embedding. Unlike candidate-based grasp generation, the diffuser predicts grasp poses directly from the shared object latent.

\subsection{Scorer and Reconstruction}
\label{sec:scorer_reconstruction}

The pose-reasoning diffuser generates continuous grasp hypotheses from the shared point-wise object latent. However, generative pose prediction alone does not determine which hypotheses are geometrically valid, locally stable, or semantically compatible with the target object. We further introduce a reconstruction-aware scorer to assess each queried surface point from three complementary aspects: graspability, grasp quality, and affordance. 

Concretely, given the fused point-wise latent $\mathbf{f}_i$, we predict three point-wise logits:
\begin{equation}
\hat{s}_i = H_{\text{grasp}}(\mathbf{f}_i), \qquad
\hat{q}_i = H_{\text{qual}}(\mathbf{f}_i), \qquad
\hat{a}_i = H_{\text{aff}}(\mathbf{f}_i).
\end{equation}
Here, $\hat{s}_i$ estimates whether the queried surface point $\mathbf{p}_i$ is graspable, $\hat{q}_i$ measures the local grasp quality around that point, and $\hat{a}_i$ captures affordance-oriented cues for manipulation. Unlike pose-only scoring, these predictions are grounded in the reconstructed object structure encoded in $\mathbf{f}_i$, which makes them suitable for both grasp ranking and reconstruction-aware latent refinement. 

Beyond grasp scoring, we further use the predicted grasp signals to refine the shared reconstruction latent. Specifically, we first compute a confidence weight for each queried point, 
\begin{equation}
\omega_i = \sigma(\hat{s}_i)\,\sigma(\hat{q}_i)\,\sigma(\hat{a}_i),
\end{equation}
and aggregate the point-wise features into a manipulation-aware summary,
\begin{equation}
\bar{\mathbf{f}} =
\frac{\sum_i \omega_i \mathbf{f}_i}{\sum_i \omega_i + \epsilon}.
\end{equation}
This summary feature is then fed to a lightweight latent updater $U_{\text{lat}}(\cdot)$ to predict a residual feature volume,
\begin{equation}
\Delta \mathbf{Z} = U_{\text{lat}}(\bar{\mathbf{f}}, \mathbf{Z}),
\end{equation}
which is added back to the original latent grid:
\begin{equation}
\mathbf{Z}' = \mathbf{Z} + \Delta \mathbf{Z}.
\end{equation}
where $\Delta \mathbf{Z}$ is produced by a lightweight latent updater conditioned on $\bar{\mathbf{f}}$. In this way, grasp supervision is written back into the reconstruction branch, so that the latent gradually captures manipulation-relevant structure in addition to geometric fidelity.

To preserve geometric consistency after latent updating, we impose a surface-aware regularization on the refined latent. For each queried point, we evaluate a surface prediction head at the surface location and at a small offset along the normal direction:
\begin{equation}
\ell_i^{+} = H_{\text{surf}}(\mathcal{S}(\mathbf{Z}', \bar{\mathbf{p}}_i)), \qquad
\ell_i^{-} = H_{\text{surf}}(\mathcal{S}(\mathbf{Z}', \bar{\mathbf{p}}_i + \delta \mathbf{n}_i)),
\end{equation}
where $\delta$ is a small scalar offset. We then optimize
\begin{equation}
\mathcal{L}_{\text{rec-upd}}
=
\frac{\sum_i \omega_i \,\mathcal{L}_{\text{bce}}(\ell_i^{+}, 1)}
{\sum_i \omega_i + \epsilon}
+
\frac{\sum_i \omega_i \,\mathcal{L}_{\text{bce}}(\ell_i^{-}, 0)}
{\sum_i \omega_i + \epsilon},
\end{equation}
where $\mathcal{L}_{\text{bce}}$ denotes the binary cross-entropy loss. This design makes the refined latent more favorable for reconstruction while aligning it with grasp-relevant affordance cues. 

\subsection{Optimization}

\paragraph{Training.}
We train GraspFoM with point-wise supervision on the queried visible surface points, using the SAM3D stage-1 latent extracted online or loaded from cache. Let $\Omega$ denote the set of valid queried points, and let $\Omega^{+}\subseteq\Omega$ denote the subset with valid positive grasp supervision. For each point $\mathbf{p}_i$, the model predicts graspness $\hat{s}_i$, quality $\hat{q}_i$, affordance $\hat{a}_i$, anchor logits $\boldsymbol{\pi}_i$, and the denoised diffusion target $\hat{\mathbf{x}}_i^0$.

For compactness, we denote binary cross-entropy and cross-entropy losses by $\ell_{\mathrm{bce}}(\cdot,\cdot)$ and $\ell_{\mathrm{ce}}(\cdot,\cdot)$, respectively. The point-wise scoring losses are: 
\begin{align}
\mathcal{L}_{\text{grasp}}
&=
\frac{1}{|\Omega|}
\sum_{i\in\Omega}
\ell_{\mathrm{bce}}(\hat{s}_i, s_i), \\
\mathcal{L}_{\text{qual}}
&=
\frac{1}{|\Omega^{+}|}
\sum_{i\in\Omega^{+}}
\lVert \hat{q}_i - q_i \rVert_2^2, \\
\mathcal{L}_{\text{aff}}
&=
\frac{1}{|\Omega|}
\sum_{i\in\Omega}
\ell_{\mathrm{bce}}(\hat{a}_i, a_i).
\end{align}

For the diffusion branch, we supervise both the denoised pose target and the anchor assignment:
\begin{align}
\mathcal{L}_{\text{diff}}
&=
\frac{1}{|\Omega^{+}|}
\sum_{i\in\Omega^{+}}
\lVert \hat{\mathbf{x}}_i^0 - \mathbf{x}_i^0 \rVert_2^2, \\
\mathcal{L}_{\text{mode}}
&=
\frac{1}{|\Omega^{+}|}
\sum_{i\in\Omega^{+}}
\ell_{\mathrm{ce}}(\boldsymbol{\pi}_i, k_i^{\star}),
\end{align}
where $k_i^{\star}$ is the index of the assigned anchor for point $i$.

When the residual latent updater is enabled, we additionally optimize the reconstruction-update loss $\mathcal{L}_{\text{rec-upd}}$ defined in Sec .~\ref {sec:scorer_reconstruction}. The overall training objective is
\begin{align}
\mathcal{L}
=
&\;\lambda_g \mathcal{L}_{\text{grasp}}
+\lambda_q \mathcal{L}_{\text{qual}}
+\lambda_a \mathcal{L}_{\text{aff}} \nonumber \\
&+\lambda_d \mathcal{L}_{\text{diff}}
+\lambda_m \mathcal{L}_{\text{mode}}
+\lambda_r \mathcal{L}_{\text{rec-upd}}.
\end{align}

In practice, the grasp-anchor bank is built once by K-means clustering over normalized valid training poses and then cached for subsequent training and inference.

\paragraph{Inference.}
At inference time, we first run the SAM3D sparse-structure stage to obtain $(\mathbf{z}, \boldsymbol{\gamma}, \boldsymbol{\beta})$, and then query visible surface points and normals from the point map. Given the fused point-wise latent, the pose-reasoning diffuser performs truncated DDIM sampling from each grasp anchor. Specifically, let $\mathbf{x}_{i,k}^{t_0}$ denote the initial noisy sample around anchor $\mathbf{a}_k$ at a truncated starting timestep $t_0$. Starting from $\mathbf{x}_{i,k}^{t_0}$, the sampler iteratively updates
\begin{equation}
\mathbf{x}_{i,k}^{t-1} = \mathrm{DDIMStep}\!\left(\hat{\mathbf{x}}_{i,k}^{0}, \mathbf{x}_{i,k}^{t}, t\right),
\end{equation}
for a small number of denoising steps, and the final denoised state is converted back to metric pose parameters by
\begin{equation}
[\hat{\mathbf{r}}_{i,k}, \hat{d}_{i,k}] = \mathcal{T}_{\text{denorm}}(\hat{\mathbf{x}}_{i,k}^{0}).
\end{equation}

The predicted pose parameters $\hat{\mathbf{r}}_{i,k}$ and $\hat{d}_{i,k}$ are further converted into full grasp poses by estimating the finger opening width from local object geometry, rather than diffusing it directly. Specifically, the width $\hat{w}_{i,k}$ is computed from the local extent of the neighborhood $\mathcal{N}(i)$ around $\mathbf{p}_i$ along the width axis induced by $\hat{\mathbf{r}}_{i,k}$, followed by clipping to the valid gripper range.

The scorer ranks sampled grasp hypotheses in two stages. We first compute a point-level score that reflects whether the queried surface point is graspable, reliable, and manipulation-relevant:
\begin{equation}
\phi_i = \sigma(\hat{s}_i)\,\hat{q}_i
\left(1 + \lambda_{\text{aff}} \sigma(\hat{a}_i)\right).
\end{equation}
Here, $\phi_i$ combines graspness, local grasp quality, and affordance cues into a reconstruction-aware point score. 

We then lift this point score to each grasp hypothesis by incorporating the view-alignment term and the anchor probability. For the grasp pose $\hat{\mathbf{g}}_{i,k}$ associated with point $i$ and anchor $k$, we define
\begin{equation}
\mathrm{Score}_{i,k} = \phi_i
\left(1 + \lambda_{\text{view}} \rho_{i,k}\right)
\,\mathrm{softmax}(\boldsymbol{\pi}_i)_k,
\end{equation}
where $\rho_{i,k}$ denotes the alignment between the predicted approach direction and the camera view ray, and $\mathrm{softmax}(\boldsymbol{\pi}_i)_k$ is the confidence of the $k$-th grasp mode. In this way, the final ranking jointly accounts for point-level graspability, geometric compatibility with the viewing condition, and the mode confidence predicted by the diffuser.

The top-ranked hypotheses are optionally filtered by collision checking. Finally, the same latent can be decoded into mesh or 3D Gaussian outputs, so the pipeline returns both grasp poses and high-fidelity 3D assets.

\section{Experiments}

\subsection{Setup}

\paragraph{Settings and metrics.}
Following prior works, we evaluate GraspFoM on two complementary tasks: grasp pose prediction and 3D object reconstruction. For grasp pose prediction, we follow the official GraspNet-1Billion benchmark and report AP, AP$_{0.8}$, and AP$_{0.4}$ on the Seen, Similar, and Novel splits. For reconstruction, we evaluate the quality of the predicted 3D object geometry on GraspNet-1B using Chamfer Distance (CD), F1-Score@10mm (F1), and Normal Consistency (NC). 

\paragraph{Datasets.}
We conduct all experiments on the GraspNet-1Billion benchmark~\cite{fang2020graspnet}. Unlike recent methods~\cite{iwase2025zerograsp} that rely on additional synthetic datasets, GraspFoM is trained using only the official GraspNet-1B training split. 
For grasp pose prediction, we evaluate on the standard Seen, Similar, and Novel benchmark splits. For reconstruction, we use the object-level 3D supervision available in GraspNet-1B to compare predicted geometry against ground-truth shapes. The ReOcS dataset collected in ZeroGrasp~\cite{iwase2025zerograsp} has not been released, which prevents us from further comparison. 

\paragraph{Baselines.}
For grasp pose prediction, we compare with representative scene-level grasping methods, including GG-CNN~\cite{morrison2018closing}, Chu\etal~\cite{chu2018real}, GPD~\cite{ten2017grasp}, PointNetGPD~\cite{liang2019pointnetgpd}, GraspNet~\cite{fang2020graspnet}, GSNet~\cite{wang2021graspness}, HGGD~\cite{chen2024efficient}, EconomicGrasp~\cite{wu2024economic}, and Ma\etal~\cite{ma2023towards}. We also compare with reconstruction-aware or unified pipelines, including CenterGrasp~\cite{chisari2024centergrasp}, ZeroGrasp~\cite{iwase2025zerograsp}, and MG-Grasp (Not Open Sourced)~\cite{wang2026mg}. For 3D reconstruction, we compare with sparse-structure baselines including MinkowskiNet~\cite{choy20194d}, OCNN~\cite{wang2020deep}, OctMAE~\cite{iwase2024zero}, and ZeroGrasp~\cite{iwase2025zerograsp}. These baselines cover direct grasp detection, geometry-aware grasping, and unified reconstruction-grasping approaches. 

\begin{table}[t]
\centering
\footnotesize
\caption{Quantitative evaluation of grasp pose prediction on the GraspNet-1B. G and R in the output columns indicate whether the method predicts grasp poses and reconstructions, respectively. Best results are shown in \textbf{bold}.}
\label{tab:graspnet_benchmark}
\setlength{\tabcolsep}{4.8pt}
\renewcommand{\arraystretch}{1.16}
\resizebox{\linewidth}{!}{
\begin{tabular}{lcc|ccc|ccc|ccc}
\toprule

\multirow{2}{*}{Method} & \multicolumn{2}{c|}{Output} & \multicolumn{3}{c|}{Seen} & \multicolumn{3}{c|}{Similar} & \multicolumn{3}{c}{Novel} \\
\cmidrule(lr){2-3} \cmidrule(lr){4-6} \cmidrule(lr){7-9} \cmidrule(lr){10-12}
 & G & R & AP & AP$_{0.8}$ & AP$_{0.4}$ & AP & AP$_{0.8}$ & AP$_{0.4}$ & AP & AP$_{0.8}$ & AP$_{0.4}$ \\
\midrule
GG-CNN~\cite{morrison2018closing} & \checkmark &  & 15.48 & 21.84 & 10.25 & 13.26 & 18.37 & 4.62 & 5.52 & 5.93 & 1.86 \\
Chu\etal~\cite{chu2018real} & \checkmark &  & 15.97 & 23.66 & 10.80 & 15.41 & 20.21 & 7.06 & 7.64 & 8.69 & 2.52 \\
CenterGrasp~\cite{chisari2024centergrasp} & \checkmark & \checkmark & 16.46 & 20.24 & 11.74 & 9.52 & 11.92 & 5.71 & 1.60 & 1.89 & 1.12 \\
GPD~\cite{ten2017grasp} & \checkmark &  & 22.87 & 28.53 & 12.84 & 21.33 & 27.83 & 9.64 & 8.24 & 8.89 & 2.67 \\
Lian\etal~\cite{liang2019pointnetgpd} & \checkmark &  & 25.96 & 33.01 & 15.37 & 22.68 & 29.15 & 10.76 & 9.23 & 9.89 & 2.74 \\
GraspNet~\cite{fang2020graspnet} & \checkmark &  & 27.56 & 33.43 & 16.59 & 26.11 & 34.18 & 14.23 & 10.55 & 11.25 & 3.98 \\
GSNet~\cite{wang2021graspness} & \checkmark &  & 67.12 & 78.46 & 60.90 & 54.81 & 66.72 & 46.17 & 24.31 & 30.52 & 14.23 \\
Ma\etal~\cite{ma2023towards} & \checkmark &  & 63.83 & 74.25 & 58.66 & 58.46 & 70.05 & 51.32 & 24.63 & 31.05 & 12.85 \\
HGGD~\cite{chen2024efficient} & \checkmark &  & 64.45 & 72.81 & 61.16 & 53.59 & 64.12 & 45.91 & 24.59 & 30.46 & 15.58 \\
EconomicGrasp~\cite{wu2024economic} & \checkmark &  & 68.21 & 79.60 & 63.54 & 61.19 & 73.60 & 53.77 & 25.48 & 31.46 & 13.85 \\
MG-Grasp~\cite{wang2026mg} & \checkmark & \checkmark & 66.80 & - & - & 57.35 & - & - & 23.22 & - & - \\
ZeroGrasp~\cite{iwase2025zerograsp} & \checkmark & \checkmark & 72.43 & 83.12 & 65.57 & 65.45 & 78.32 & 55.48 & 28.49 & 34.21 & 15.80 \\
\midrule
\textbf{GraspFoM (Ours)} & \checkmark & \checkmark & \best{78.87} & \best{89.60} & \best{71.13} & \best{73.34} & \best{87.84} & \best{64.12} & \best{54.32} & \best{64.01} & \best{29.36} \\
\bottomrule
\end{tabular}
}
\end{table}

\begin{table}[t]
\centering
\begin{minipage}[t]{0.42\linewidth}
\centering
\vspace{1.5cm}
\scriptsize
\caption{3D reconstruction on GraspNet-1B. Seg. indicates segmented reconstruction. Best results are in \textbf{bold}.}
\label{tab:main_result_reconstruction}
\setlength{\tabcolsep}{3.4pt}
\renewcommand{\arraystretch}{1.18}
\resizebox{0.98\linewidth}{!}{
\begin{tabular}{lc|ccc}
\toprule
\multirow{2}{*}{Method} 
& \multirow{2}{*}{\shortstack[c]{Segmented\\Reconstruction}} 
& \multicolumn{3}{c}{GraspNet-1B} \\
\cmidrule(lr){3-5}
& & CD$\downarrow$ & F1$\uparrow$ & NC$\uparrow$ \\
\midrule
Minkowski~\cite{choy20194d} & \checkmark & 6.84 & 81.45 & 77.89 \\
OCNN~\cite{wang2020deep} & \checkmark & 7.23 & 82.22 & 78.44 \\
OctMAE~\cite{iwase2024zero} &  & 7.57 & 78.38 & 75.19 \\
ZeroGrasp~\cite{iwase2025zerograsp} & \checkmark & 6.05 & 84.08 & 78.46 \\
\midrule
\textbf{GraspFoM (Ours)} & \checkmark & \best{2.74} & \best{96.08} & \best{87.18} \\
\bottomrule
\end{tabular}
}
\vspace{0.35em}
\begin{minipage}{0.95\linewidth}
\footnotesize

\end{minipage}
\end{minipage}
\hfill
\begin{minipage}[t]{0.55\linewidth}
\centering
\scriptsize
\setlength\tabcolsep{3.1pt}
\renewcommand\arraystretch{0.94}
\caption{\textbf{Ablation study.} We evaluate anchor number, diffusion horizon, refinement steps, and reconstruction update. Light green rows denote default settings.}
\label{tab:ablation_all}
\resizebox{\linewidth}{!}{
\begin{tabular}{p{0.18\linewidth} p{0.32\linewidth}|ccc}
\toprule
\multicolumn{2}{l|}{\textbf{Grasp Parameter Variants}} 
& \makecell{\textbf{Seen}\\ AP$\uparrow$}
& \makecell{\textbf{Similar}\\ AP$\uparrow$}
& \makecell{\textbf{Novel}\\ AP$\uparrow$} \\
\midrule
\rowcolor{oursgreen}
\multicolumn{2}{l|}{\textbf{Default setting} ($K=8,\ T_{\max}=50,\ N_{\mathrm{ref}}=4$)}
& \textbf{78.87} & \textbf{73.34} & \textbf{54.32} \\
\cmidrule(lr){1-5}
\multicolumn{2}{l|}{\textit{Anchor Number} ($K$)} & & & \\
\multicolumn{2}{l|}{$K=4$} & 75.12 & 68.45 & 48.60 \\
\multicolumn{2}{l|}{$K=12$} & 77.05 & 71.12 & 51.95 \\
\cmidrule(lr){1-5}
\multicolumn{2}{l|}{\textit{Diffusion Horizon} ($T_{\max}$)} & & & \\
\multicolumn{2}{l|}{$T_{\max}=40$} & 74.58 & 67.90 & 47.85 \\
\multicolumn{2}{l|}{$T_{\max}=60$} & 77.40 & 71.85 & 52.10 \\
\cmidrule(lr){1-5}
\multicolumn{2}{l|}{\textit{Refinement Steps} ($N_{\mathrm{ref}}$)} & & & \\
\multicolumn{2}{l|}{$N_{\mathrm{ref}}=2$} & 73.20 & 65.50 & 45.10 \\
\multicolumn{2}{l|}{$N_{\mathrm{ref}}=6$} & 77.65 & 72.05 & 52.80 \\

\specialrule{0.6pt}{0.08em}{0.08em}

\multicolumn{2}{l|}{\textbf{Reconstruction Ablation}} 
& \textbf{CD}$\downarrow$
& \textbf{F1}$\uparrow$
& \textbf{NC}$\uparrow$ \\
\midrule
\multicolumn{2}{l|}{\textit{Reconstruction Update}} & & & \\
\multicolumn{2}{l|}{w/o Reconstruction Update} & 3.58 & 91.25 & 80.45 \\
\rowcolor{oursgreen}
\multicolumn{2}{l|}{Full GraspFoM} & \textbf{2.74} & \textbf{96.08} & \textbf{87.18} \\
\bottomrule
\end{tabular}
}
\end{minipage}
\end{table}


Due to space limitations, we defer model details and training details to the supplementary material. \textbf{Please refer to the appendix for additional experiments and visualization results.} 

\begin{figure*}[!ht]
  \includegraphics[width=\textwidth]{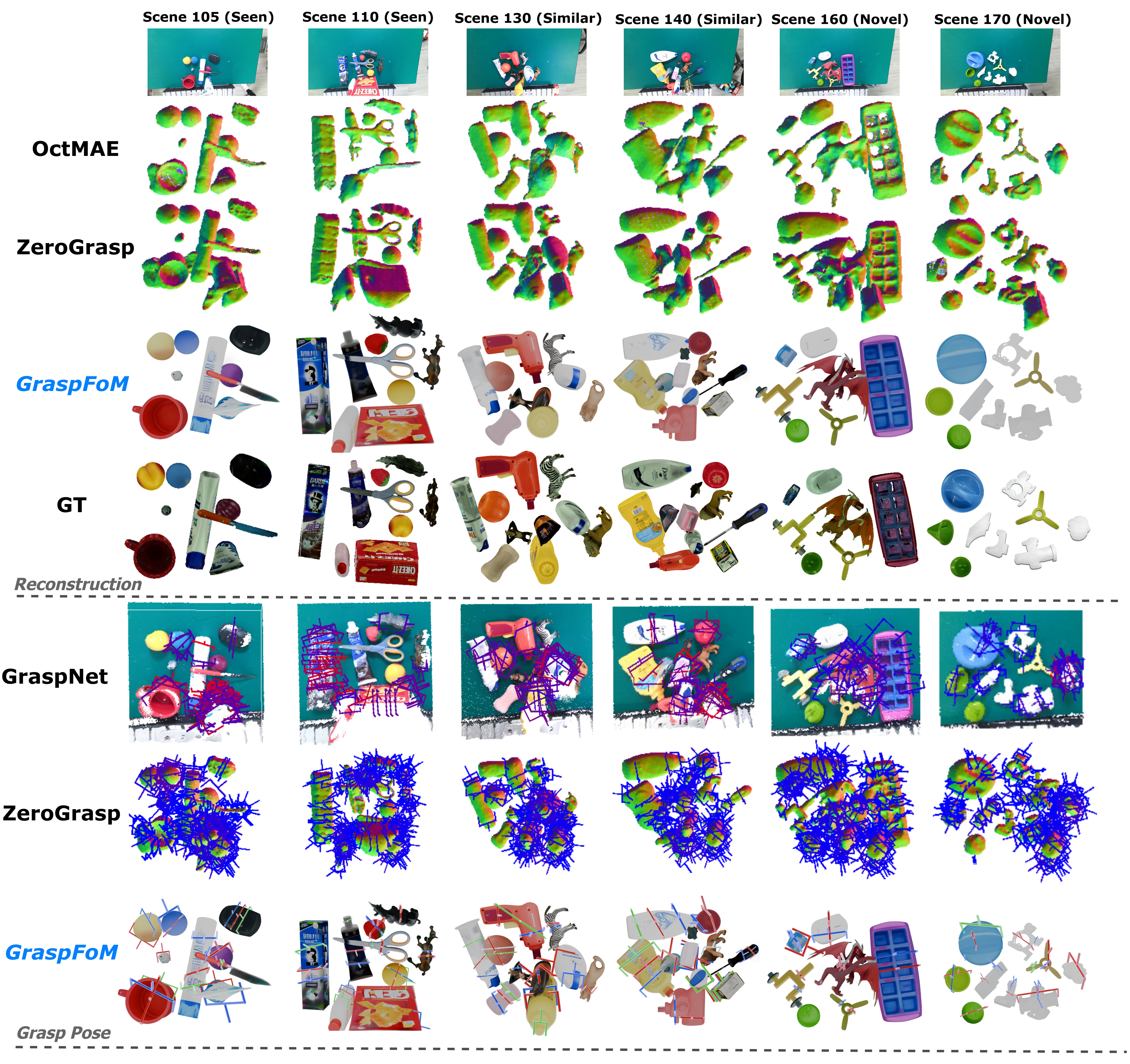}
  \vspace{-6mm}
  \caption{Reconstruction and grasping pose visualization comparison.}
  \vspace{-4mm}
  \label{fig:reconstruction_vis}
\end{figure*}

\begin{figure}[t]
  \centering
  \includegraphics[width=\linewidth]{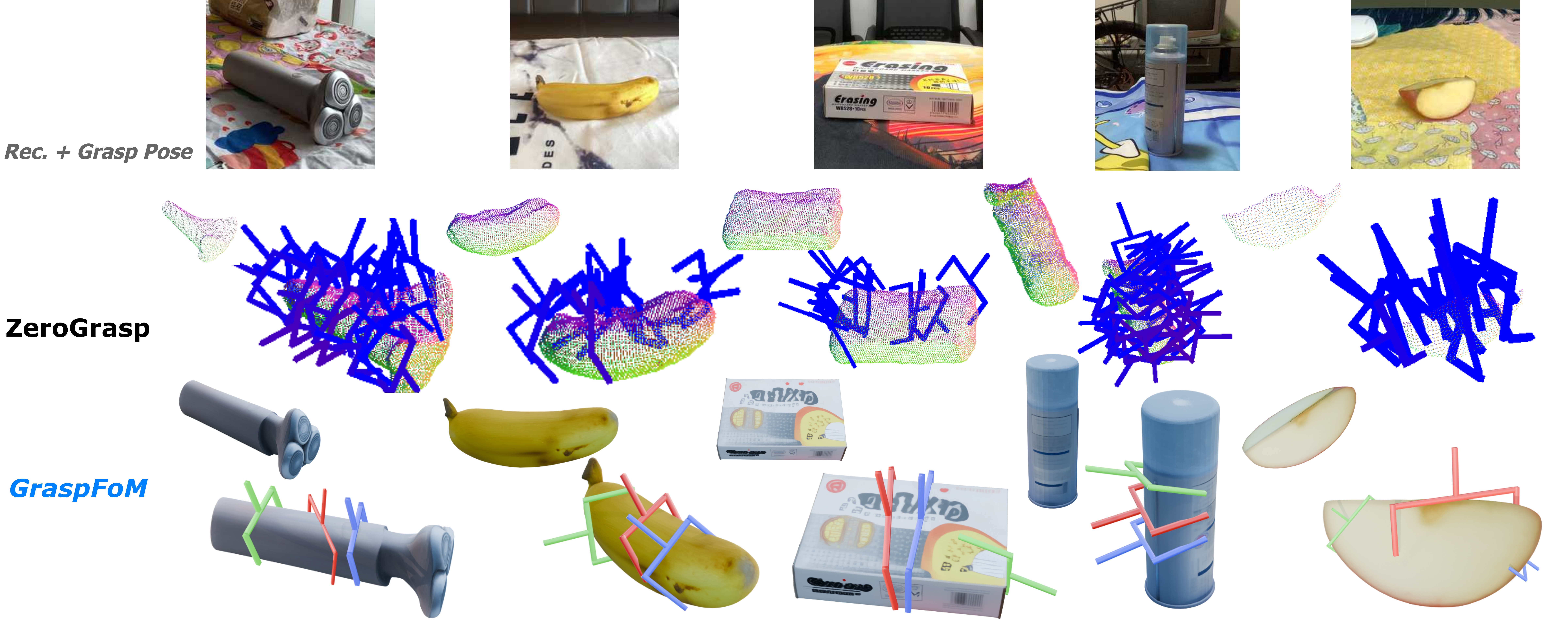}
  \vspace{-2mm}
  \caption{In-the-wild reconstruction and grasp pose visualization comparison.}
  \label{fig:wild_vis}
\end{figure}

\subsection{Main Results}

\paragraph{Grasp Pose Prediction.}
Tab.~\ref{tab:graspnet_benchmark} reports quantitative results on the GraspNet-1B benchmark. GraspFoM achieves the strongest overall performance across all benchmark splits and all reported friction settings. More importantly, the gain is especially pronounced on the Novel split, which is the most challenging setting because the object categories are not observed during training. This result suggests that the learned representation is not merely fitting training-set grasp patterns, but instead transfers to previously unseen objects through reusable object-level 3D priors. 

Existing reconstruction-aware grasping methods typically learn task-specific latent spaces from scratch, whereas GraspFoM starts from transferable SAM3D priors and further builds shape-point fused features that combine local surface evidence with global shape context. This design is particularly beneficial under occlusion and partial observations, where purely local point neighborhoods are often insufficient to determine stable grasps. The strong improvement on novel objects supports this view. 

Fig.~\ref{fig:reconstruction_vis} leads to the same conclusion. GraspFoM produces more reliable grasp poses with better contact regions and more stable approach directions. We attribute this behavior to the shared object latent, which jointly captures point-level local geometry and object-level global structure. As a result, the model can reason about graspable regions beyond visible local fragments and better preserve the overall object layout when predicting grasps. 

Another notable aspect of this comparison is the training protocol. ZeroGrasp reports a stronger setting by pretraining on its additional ZeroGrasp-11B dataset and then fine-tuning on GraspNet-1B, while MG-Grasp uses sparse multi-view RGB observations with known camera poses and a different input setup. By contrast, GraspFoM is trained only on the official GraspNet-1B training split. The strong performance indicates that foundation priors can provide a more data-efficient route to robust grasp generalization.

\paragraph{3D Reconstruction.}
Tab.~\ref{tab:main_result_reconstruction} reports reconstruction performance on GraspNet-1B. GraspFoM also achieves the best overall geometry quality, demonstrating that the proposed framework improves not only grasp prediction but also object reconstruction itself. This result is important because our method does not treat reconstruction merely as an auxiliary branch. Instead, reconstruction and grasping are optimized jointly through the shared latent, the reconstruction-aware scorer, and the residual latent updater.

Compared with sparse-structure baselines such as MinkowskiNet, OCNN, and OctMAE, GraspFoM benefits from stronger object priors and a reconstruction process that is explicitly shaped by grasp supervision. Compared with ZeroGrasp, which also performs joint reconstruction and grasp prediction, our method is built on top of transferable 3D foundation priors rather than a task-specific latent space learned from scratch. As a result, the predicted geometry is more favorable for both accurate surface recovery and downstream grasp reasoning.

Another notable observation is that the reconstruction improvement is consistent with the gain in grasp prediction. This suggests that the learned latent is not simply better for one task at the expense of the other. Instead, the two tasks reinforce each other: better reconstruction provides more grounded geometric support for grasp generation, while grasp supervision encourages the latent to capture manipulation-relevant structure, such as graspable regions and affordance-aware cues. This mutual reinforcement is a key reason why GraspFoM is able to outperform prior unified frameworks on both tasks simultaneously.


\subsection{Ablation Study}

We conduct ablation studies on GraspNet-1B to examine the contributions of several key design choices in GraspFoM, including the anchor number $K$, the diffusion horizon $T_{\max}$, the number of refinement steps $N_{\mathrm{ref}}$, and the reconstruction update. For the grasp generation branch, we report AP on the Seen, Similar, and Novel splits. For the reconstruction branch, we report CD, F1, and NC. Unless otherwise specified, the default configuration uses $K=8$, $T_{\max}=50$, and $N_{\mathrm{ref}}=4$, which is highlighted in light green in Tab.~\ref{tab:ablation_all}. 

The results show that all three hyperparameters have a clear impact on grasp prediction performance. For anchor number, using $K=8$ gives the best AP across all three splits, outperforming both a smaller setting ($K=4$) and a larger one ($K=12$). This suggests that too few anchors limit the diversity of candidate grasp hypotheses, while too many anchors introduce redundancy and make optimization less effective. A similar trend is observed for the diffusion horizon. Setting $T_{\max}=50$ achieves the best results, whereas a shorter horizon ($T_{\max}=40$) leads to clear degradation, indicating insufficient denoising capacity, and a longer horizon ($T_{\max}=60$) also slightly hurts performance, suggesting diminishing returns from overly long diffusion trajectories. For iterative refinement, $N_{\mathrm{ref}}=4$ performs best. Reducing the number of refinement steps to $2$ significantly weakens performance, especially on the Novel split, while increasing it to $6$ does not bring further gains. This indicates that a moderate refinement budget is sufficient to capture most of the benefit, while excessive refinement yields limited improvement.

We further ablate the reconstruction update to verify the role of reconstruction-aware learning in GraspFoM. Removing this module degrades reconstruction quality from 2.74 to 3.58 in CD, from 96.08 to 91.25 in F1, and from 87.18 to 80.45 in NC. These results show that the reconstruction update is important for learning more accurate and geometrically consistent object representations. In turn, this stronger 3D representation also benefits grasp generation, supporting our core design that grasp reasoning should be built on top of high-quality object-centric 3D priors rather than only partial observations.

\subsection{In-the-wild Evaluation}

We further conduct qualitative evaluation on several object scenes selected from the WildRGB-D~\cite{xia2024rgbd} dataset, using only the RGB images as input, as shown in Fig.~\ref{fig:wild_vis}. Compared with ZeroGrasp, GraspFoM produces more complete reconstructions and stable grasp poses. This advantage comes from the shared object latent, which combines local point cues with global shape context from 3D foundation priors for more robust reconstruction and grasp reasoning in open-world scenes. 

\section{Conclusion}

In this paper, we present GraspFoM, a novel unified framework for robotic grasping with 3D foundation priors. We form a shared 3D object latent that jointly supports object reconstruction and grasp pose prediction. GraspFoM achieves SOTA performance on GraspNet-1B and simultaneously produces high-fidelity 3D assets. 

\paragraph{Limitations and Future Work.}
The current framework mainly focuses on rigid objects. Incorporating structured kinematic priors, such as URDF~\cite{li2025urdf,le2024articulate}, is a possible direction for articulated-object manipulation.


\bibliographystyle{corlabbrvnat}
\bibliography{paper}

\clearpage
\appendix

\section{Supplementary Material}

\subsection{Overview}
The supplementary material includes the following components:
\begin{itemize}
    \item Data Preprocessing
    \item Training Details
    \item Implementation Details
    \item Additional Visualization Results
\end{itemize}

\subsection{Data Preprocessing}
Our pipeline is built on top of object-centric outputs from the SAM3D reconstruction. For each training step, we use the RGB image together with the object mask to obtain an object-aligned 3D representation, including a pointmap and a compact volumetric shape latent. To reduce training overhead, these intermediate outputs are cached offline and reused during grasp training. As a result, the grasp model is trained on shared 3D object latents rather than repeatedly invoking the reconstruction backbone online.

For grasp supervision, we use GraspNet-1B annotations and associate them with object-level query points. In the current implementation, the query points are expressed in the camera coordinate frame. To control memory and improve training stability, the raw object point cloud is further reduced by voxel-based downsampling, with a default voxel size of $5$\,mm.

The grasp pose labels are converted into a diffusion-friendly representation before training. Specifically, each grasp pose is parameterized by a 6D rotation representation and a grasp-depth term, which are then normalized into a bounded target space. The finger opening is not directly included in the diffusion target; instead, it is recovered later from local object geometry during inference. To model multimodal pose structure, we additionally construct a compact anchor set by collecting normalized ground-truth pose targets from the training set and applying offline $k$-means clustering. These anchors serve as coarse pose modes for subsequent classification and initialization. 

\subsection{Implementation Details}

\paragraph{Architecture.} Our grasp branch is trained on top of cached SAM3D pointmaps and shape latents, so that all predictions are built on a shared 3D object representation. The shared latent is represented as an $8$-channel volumetric grid with resolution $16 \times 16 \times 16$. For each query point, we concatenate its 3D coordinates and surface normal as input to a lightweight point encoder. Local latent features are obtained by trilinear sampling from the latent grid, while global shape context is introduced through cross-attention with latent tokens. The resulting fused point token is used for graspness, quality, affordance, and pose prediction.

\paragraph{Training.} For pose learning, we adopt anchor-initialized truncated diffusion. Each positive query point is assigned to its nearest anchor in the normalized diffusion space, which provides supervision for anchor classification. Given a clean pose target $x_0$, we sample a truncated timestep and generate a noisy state $x_t$, and train the pose reasoner to recover $x_0$ from $(x_t,t)$. The overall objective is a weighted combination of graspness, grasp quality, affordance, anchor classification, and diffusion denoising losses. The implementation additionally supports an optional reconstruction-aware latent refinement branch, denoted as \textit{update\_Reconstruction}, with an auxiliary surface consistency loss.

\paragraph{Inference.} Optimization is performed with AdamW for $20$ epochs using a learning rate of $10^{-4}$ and a batch size of $32$. Training is conducted with distributed data parallelism on $8$ GPUs. At inference time, grasp generation starts from the selected anchor with Gaussian perturbation and is refined by truncated DDIM. Unless otherwise stated, the default setup uses a diffusion horizon of $50$, initializes reverse denoising from $t_{\mathrm{init}}=16$, and performs $5$ refinement steps. The finger opening is estimated separately from local object geometry along the predicted gripper width axis, rather than being directly generated by the diffusion branch.

\subsection{Additional Visualization Results}
\label{sec:supp_vis_occlusion}

\begin{figure}[t]
    \centering
    \includegraphics[width=0.5\linewidth]{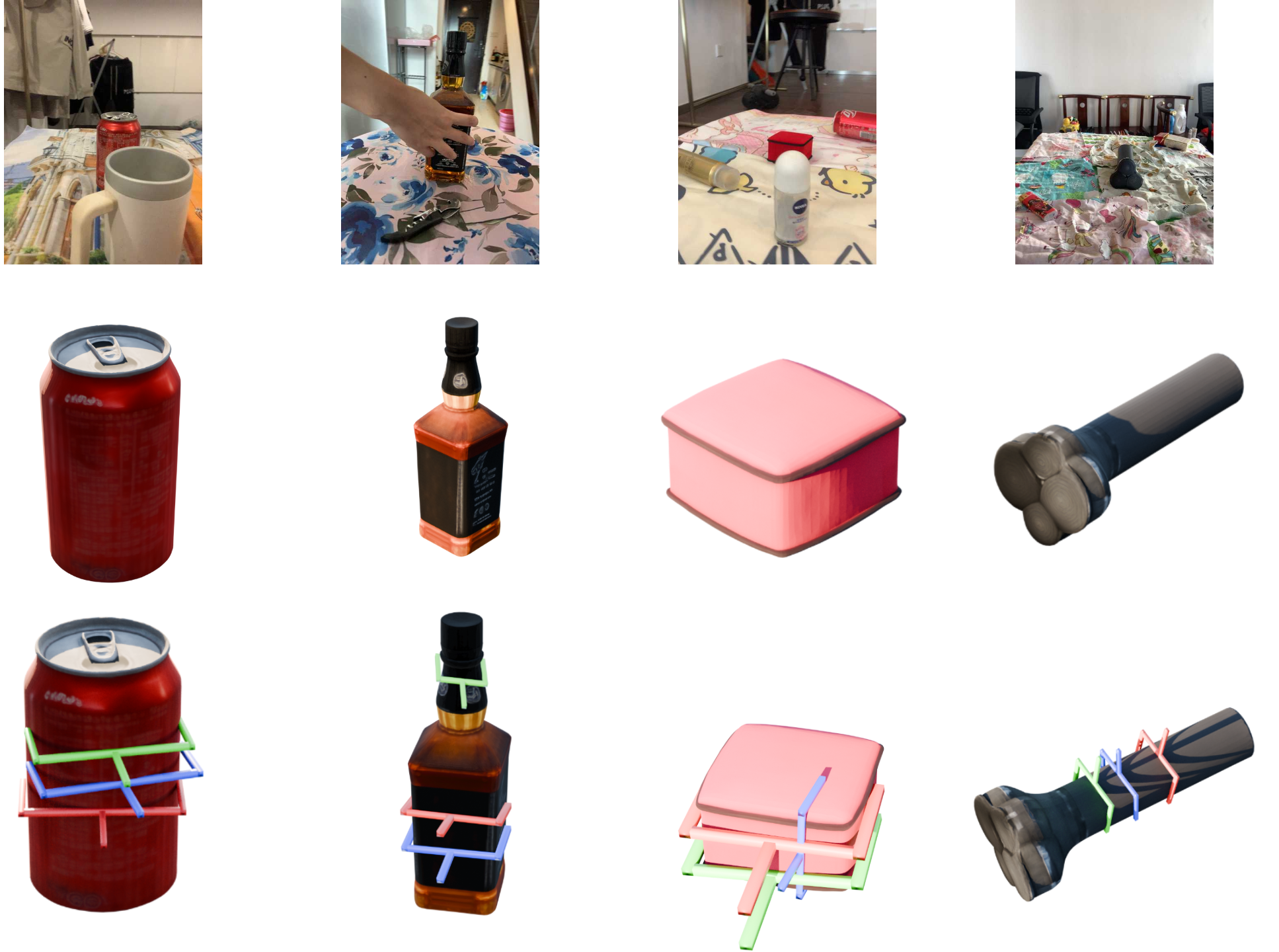}
    \caption{Additional qualitative results on occluded and poor-view objects.}
    \label{fig:supp_occlusion_vis}
\end{figure}

We further provide qualitative results on several challenging object scenes with severe occlusion or unfavorable viewpoints, as shown in Fig.~\ref{fig:supp_occlusion_vis}.

These examples further verify the robustness of GraspFoM under partial observations. Even when only limited visible evidence is available, our method still reconstructs geometrically coherent object shapes and predicts stable grasp poses with reasonable contact regions and approach directions. This advantage comes from the shared object latent, which combines local point cues with global shape context from transferable 3D foundation priors, making grasp reasoning less dependent on directly visible local fragments.
As a result, GraspFoM not only produces more complete reconstructions but also yields grasps that are better aligned with the recovered object structure.

\subsection{Additional Material Description}
In the supplementary folder, we additionally provide representative case materials, including the input RGB images, corresponding output results, and mesh files for qualitative visualization. We also include a compact release of the core method code for presentation and discussion, covering the main modules of point-shape fusion, pose reasoning, and diffusion-based grasp refinement, so that the overall input-output pipeline and the key implementation of our method can be more clearly understood.

\end{document}